\documentclass{article}

\usepackage{arxiv}
\usepackage[utf8]{inputenc} 
\usepackage[T1]{fontenc}    
\usepackage{hyperref}       
\usepackage{url}            
\usepackage{booktabs}       
\usepackage{amsfonts}       
\usepackage{nicefrac}       
\usepackage{microtype}      
\usepackage{mathtools}
\usepackage{nccmath}
\usepackage{physics}
\usepackage{cleveref}       
\usepackage{graphicx}
\usepackage{natbib}
\usepackage{doi}

\DeclarePairedDelimiter{\nint}\lfloor\rceil
\newcommand\U{\textrm{U}}
\newcommand\BIC{\textrm{BIC}}
\newcommand\UBIC{\textrm{UBIC}}
\newcommand\ICOMP{\textrm{ICOMP}}

\title{On uncertainty-penalized Bayesian information criterion}

\date{}

\author{\href{https://orcid.org/0000-0001-7991-315X}{\includegraphics[scale=0.06]{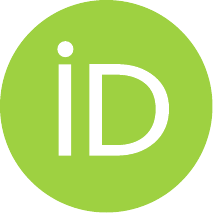}\hspace{1mm}Pongpisit Thanasutives}\thanks{Corresponding author} \\
	Graduate School of Information Science and Technology\\
	Osaka University\\
	Osaka, Japan \\
	\texttt{thanasutives@ai.sanken.osaka-u.ac.jp} \\
	\And
	\href{https://orcid.org/0000-0002-2451-1919}{\includegraphics[scale=0.06]{orcid.pdf}\hspace{1mm}Ken-ichi Fukui} \\
	SANKEN (The Institute of Scientific and Industrial Research)\\
	Osaka University\\
	Osaka, Japan \\
	\texttt{fukui@ai.sanken.osaka-u.ac.jp} \\
}

\renewcommand{\headeright}{}
\renewcommand{\undertitle}{}
\renewcommand{\shorttitle}{On uncertainty-penalized Bayesian information criterion}

\begin{document}
\maketitle

\begin{abstract}
	The uncertainty-penalized information criterion (UBIC) has been proposed as a new model-selection criterion for data-driven partial differential equation (PDE) discovery. In this paper, we show that using the UBIC is equivalent to employing the conventional BIC to a set of overparameterized models derived from the potential regression models of different complexity measures. The result indicates that the asymptotic property of the UBIC and BIC holds indifferently.
\end{abstract}

\keywords{Bayesian information criterion \and data-driven partial differential equation discovery \and uncertainty quantification}

\section{Definitions}
Suppose governing nonlinear dynamics of a physical system is identifiable through approximating the state variable's time derivative $u_{t}$ using a candidate library $\Phi$ (containing $N$ samples) and a vector of coefficients $\xi$. The conventional BIC \citep{BIC} reads as 

\begin{equation} \label{eq:BIC}
	\BIC_{\Phi}(\xi) = -2\log\mathcal{L}(\Phi, \xi)  + \log(N)\norm{\xi}_{0};
\end{equation}

where $\mathcal{L}(\Phi, \xi)$ is the maximized likelihood function value of the model, parameterized by $\xi$ and making the approximation on $\Phi$. According to \citep{thanasutives2024adaptive}, the UBIC penalizes a model or PDE by its quantified uncertainty $\U_{\xi}$ to prevent the selection of an overfitted PDE with unimportant terms given by the BIC \citep{nPIML}. The criterion is defined as follows:

\begin{equation} \label{eq:UBIC}
	\UBIC_{\Phi}(\xi, \U_{\xi}) = -2\log\mathcal{L}(\Phi, \xi)  + \log(N)(\norm{\xi}_{0} + \U_{\xi}).
\end{equation}

\section{Main proof (Connecting the BIC and UBIC)}
Given the quantified uncertainty $\U_{\xi}$ of every potential model, each representing a PDE with a specific number of nonzero terms (support size), we have an approximate of the time derivative

\begin{equation} \label{eq:old_estimation}
	\hat{u}_{t} = \Phi\xi + \epsilon;\, \norm{\xi}_{0}=k,\, p=k+1.
\end{equation}

Here $p$ tells the total degree of freedom including an intercept $\epsilon$. We can also write Equation (\ref{eq:old_estimation}) in the compact form $\hat{u}_{t} = \Phi^{\prime}\xi^{\prime}$ ($\norm{\xi^{\prime}}_{0} = p$) without explicitly showing $+\epsilon$. Our objective is to find the corresponding $\Tilde{\Phi}$ and $\hat{\xi}$ such that 

\begin{equation} \label{eq:new_estimation}
\begin{split}
	\hat{u}_{t} &= \Tilde{\Phi}\hat{\xi};\, \norm{\hat{\xi}}_{0}=\U_{\xi}+p, \\
	\Tilde{\Phi} &= \begin{pmatrix} \Phi &\bigm| & \Tilde{A} \end{pmatrix},\, \textrm{and}\, 
	\hat{\xi} = \begin{pmatrix} \xi \\ \Tilde{\xi} \end{pmatrix}.
\end{split}
\end{equation}

Equation (\ref{eq:new_estimation}) partitions the coefficient vector $\hat{\xi}$ into two components. The first component is expected to encompass certain true terms, and the second component comprises overparameterized coefficients introduced by $\U_{\xi}$. Note that although $\U_{\xi}$ is generally a positive real number (greater than or equal to $1$) bounded from above, we can always round it to be an integer, i.e. $\U_{\xi} \leftarrow \nint{\U_{\xi}}$. As a consequence, we have $\epsilon = \Tilde{A}\Tilde{\xi}$ and $\norm{\Tilde{\xi}}_{0} = \U_{\xi} + 1 \geq 2$. if there exists such a pair of $\Tilde{\Phi}$ and $\hat{\xi}$, we essentially find a new set of overparameterized models, on which we compute the BIC, to match the UBIC values obtained using Equation (\ref{eq:UBIC}). Since we do not assume $\Tilde{A}$ to be linearly dependent, we can, for example, set 

\begin{equation}
	\epsilon = \Tilde{A}\Tilde{\xi} = 
	\begin{pmatrix}
		\frac{\epsilon}{\U_{\xi} + 1} & \cdots & \frac{\epsilon}{\U_{\xi} + 1}
	\end{pmatrix}
	\begin{pmatrix}
		1 \\ \vdots \\ 1
	\end{pmatrix};
\end{equation}

where all elements in $\Tilde{\xi}$ are nonzero. Because the time derivative approximation is identical in both Equation (\ref{eq:old_estimation}) and (\ref{eq:new_estimation}), we consequently have $\log\mathcal{L}(\Tilde{\Phi}, \hat{\xi}) = \log\mathcal{L}(\Phi^{\prime}, \xi^{\prime})$. Our main claim is now achieved by deriving

\begin{equation}
\begin{split}
	\UBIC_{\Phi^{\prime}}(\xi^{\prime}, \U_{\xi}) &= -2\log\mathcal{L}(\Phi^{\prime}, \xi^{\prime})  + \log(N)(\norm{\xi^{\prime}}_{0} + \U_{\xi})\\ 
	&= -2\log\mathcal{L}(\Tilde{\Phi}, \hat{\xi}) + \log(N)(\norm{\hat{\xi}}_{0})\\ 
	&= \BIC_{\Tilde{\Phi}}(\hat{\xi}).
\end{split}
\end{equation}

The result essentially signifies that it is valid to view the UBIC as the BIC in the special case computed on the overparameterized vector of coefficients $\hat{\xi}$ coupled with $\Tilde{\Phi}$. We therefore conclude that when $N \rightarrow \infty$, the asymptotic property of the UBIC and BIC also holds indifferently.

\section{Necessity of a new physics-inspired model-selection criterion for identifying governing equations}
Empirical evidence shows that the UBIC can correctly identify governing equations, as the governing equations typically exhibit the lowest quantified uncertainty among potential models. However, if this observation does not hold, meaning that the quantified uncertainty $\U_{\xi}$ is a monotonically increasing function of the number of nonzero terms $\norm{\xi}_{0}$, we would not have the informative uncertainty-penalized complexity term anymore. Now the crucial question becomes: \textit{How can we identify governing equations using an information criterion with a complexity term that monotonically increases with the number of nonzero terms?}

In order to address the question, we revisit the classic experiment on the data-driven discovery of the governing Burgers' PDE (refer to the first example discussed in \citep{thanasutives2024adaptive}). However, this time, we use the informational complexity criterion (ICOMP) \citep{ICOMP} with a parametric complexity term to select the optimal support size and model as the governing equation. We consider the following ICOMP formulation based on the estimated inverse-Fisher information matrix (IFIM).

\begin{equation} \label{eq:ICOMP}
	\ICOMP_{\Phi}(\xi) = -2\log\mathcal{L}(\Phi, \xi)  + 2a_{N}C_{\xi}(\hat{\mathfrak{F}}^{-1});
\end{equation}

Here $C_{\xi}(\hat{\mathfrak{F}}^{-1})$ is the maximal information complexity of $\hat{\mathfrak{F}}^{-1}$, the estimated IFIM which depends on $\xi$. $a_{N}$ is a sequence of positive numbers, controlling the complexity penalty and affecting the ICOMP's asymptotic properties \citep{ICOMP}.

\begin{figure}
	\center
	\includegraphics{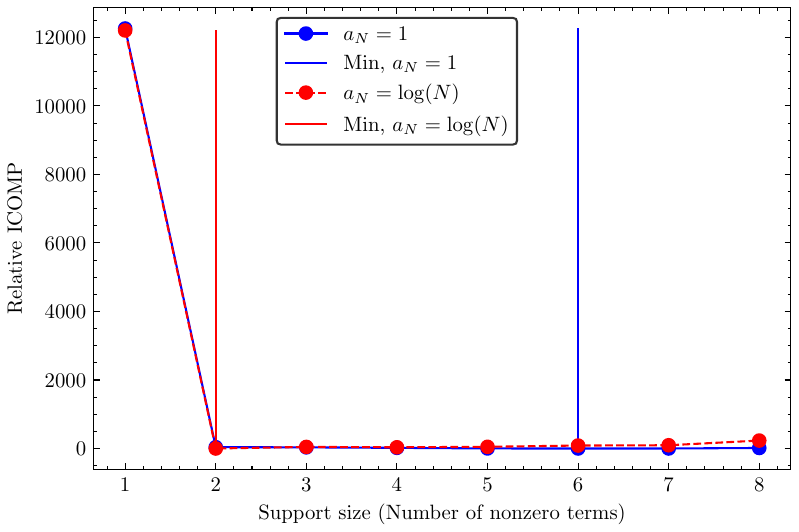}
	\caption{Plot of the relative ICOMP with respect to $\norm{\xi}_{0}$. The relative ICOMP is defined as $\ICOMP_{\Phi}(\xi) - \min_{\xi} \ICOMP_{\Phi}(\xi)$. The code implementation of the ICOMP is provided in \url{https://github.com/Pongpisit-Thanasutives/ICOMP}.}
	\label{fig:icomp}
\end{figure}

\begin{figure}
	\center
	\includegraphics{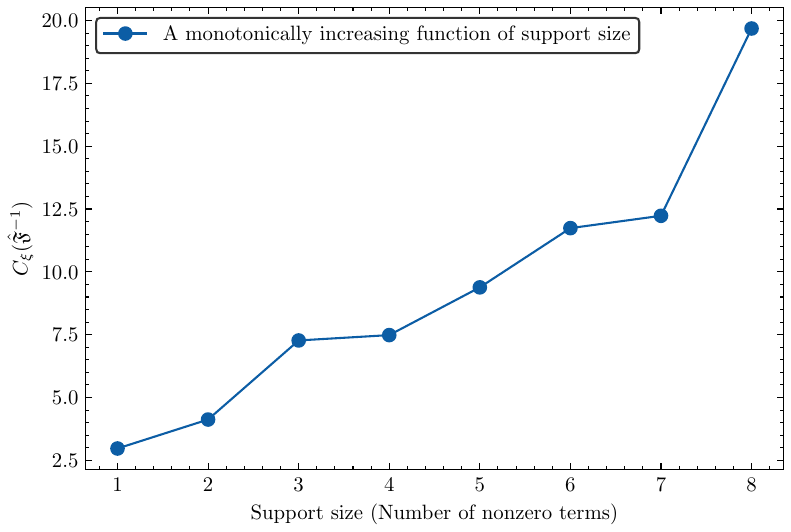}
	\caption{Plot of $C_{\xi}(\hat{\mathfrak{F}}^{-1})$ (the maximal information complexity of the estimated IFIM) with respect to $\norm{\xi}_{0}$.}
	\label{fig:ifim}
\end{figure}

In Figure \ref{fig:icomp}, we plot the model-selection results for the two common settings: $a_{N} = 1$ and $a_{N} = \log(N)$. We clearly see that the ICOMP with the complexity penalty $C_{\xi}(\hat{\mathfrak{F}}^{-1})$, sufficiently increased to $a_{N} = \log(N) = \log 10000 \approx 9.21$, can identify the true Burgers' PDE form, although $C_{\xi}(\hat{\mathfrak{F}}^{-1})$ monotonically increases by the number of nonzero terms, as seen in Figure \ref{fig:ifim}. The experimental finding sheds light on answering our question by automatically optimizing $a_{n}$ such that $2a_{N}C_{\xi}(\hat{\mathfrak{F}}^{-1})$ is large enough for identifying the governing equation. Ultimately, we posit that this newly found concept could motivate the development of a new physics-inspired model-selection criterion that benefits every PDE discovery method.

\bibliographystyle{unsrtnat}
\bibliography{ref}

\begin{thebibliography}{4}
\providecommand{\natexlab}[1]{#1}
\providecommand{\url}[1]{\texttt{#1}}
\expandafter\ifx\csname urlstyle\endcsname\relax
  \providecommand{\doi}[1]{doi: #1}\else
  \providecommand{\doi}{doi: \begingroup \urlstyle{rm}\Url}\fi

\bibitem[Schwarz(1978)]{BIC}
Gideon Schwarz.
\newblock Estimating the dimension of a model.
\newblock \emph{The annals of statistics}, pages 461--464, 1978.

\bibitem[Thanasutives et~al.(2024)Thanasutives, Morita, Numao, and Fukui]{thanasutives2024adaptive}
Pongpisit Thanasutives, Takashi Morita, Masayuki Numao, and Ken-ichi Fukui.
\newblock Adaptive uncertainty-penalized model selection for data-driven pde discovery.
\newblock \emph{IEEE Access}, 12:\penalty0 13165--13182, 2024.
\newblock \doi{10.1109/ACCESS.2024.3354819}.

\bibitem[Thanasutives et~al.(2023)Thanasutives, Morita, Numao, and Fukui]{nPIML}
Pongpisit Thanasutives, Takashi Morita, Masayuki Numao, and Ken-ichi Fukui.
\newblock Noise-aware physics-informed machine learning for robust pde discovery.
\newblock \emph{Machine Learning: Science and Technology}, 4\penalty0 (1):\penalty0 015009, feb 2023.
\newblock \doi{10.1088/2632-2153/acb1f0}.
\newblock URL \url{https://dx.doi.org/10.1088/2632-2153/acb1f0}.

\bibitem[Bozdogan and Haughton(1998)]{ICOMP}
Hamparsum Bozdogan and Dominique~M.A. Haughton.
\newblock Informational complexity criteria for regression models.
\newblock \emph{Computational Statistics \& Data Analysis}, 28\penalty0 (1):\penalty0 51--76, 1998.
\newblock ISSN 0167-9473.
\newblock \doi{https://doi.org/10.1016/S0167-9473(98)00025-5}.
\newblock URL \url{https://www.sciencedirect.com/science/article/pii/S0167947398000255}.

\end{thebibliography}

\end{document}